\documentclass{article}

\usepackage{PRIMEarxiv}

\usepackage{tabularx}
\usepackage{enumitem}

\usepackage{amsmath}
\usepackage{amssymb}
\usepackage{mathtools}
\usepackage{amsthm}
\theoremstyle{plain}

\theoremstyle{definition}

\theoremstyle{remark}

\usepackage[utf8]{inputenc} 
\usepackage[T1]{fontenc}    
\usepackage{hyperref}       
\usepackage{url}            
\usepackage{booktabs}       
\usepackage{amsfonts}       
\usepackage{nicefrac}       
\usepackage{microtype}      
\usepackage{lipsum}
\usepackage{fancyhdr}       
\usepackage{graphicx}       
\graphicspath{{media/}}     

\title{Scalable Hybrid Classification-Regression Solution for High-Frequency Nonintrusive Load Monitoring
}

\author{
  Govind Saraswat \\
  National Renewable Energy Laboratory \\
  Golden, CO, USA  \\
  \texttt{govind.saraswat@nrel.gov} \\
   \And
  Blake Lundstrom \\
  Enphase \\
  Austin, TX, USA  \\
  \texttt{blakelundstrom@gmail.com} \\
   \And
  Murti V Salapaka \\
  University of Minnesota \\
  Minneapolis, MN, USA  \\
  \texttt{murtis@umn.edu} \\
}

\begin{document}
\maketitle

\begin{abstract}
Residential buildings with the ability to monitor and control their net-load (sum of load and generation) can provide valuable flexibility to power grid operators. We present a novel multiclass nonintrusive load monitoring (NILM) approach that enables effective net-load monitoring capabilities at high-frequency with minimal additional equipment and cost. The proposed machine learning based solution provides accurate multiclass state predictions while operating at a faster timescale (able to provide a prediction for each 60-Hz ac cycle used in US power grid) without relying on event-detection techniques. We also introduce an innovative hybrid classification-regression method that allows for the prediction of not only load on/off states via classification but also individual load operating power levels via regression. A test bed with eight residential appliances is used for validating the NILM approach. Results show that the overall method has high accuracy and, good scaling and generalization properties. Furthermore, the method is shown to have sufficient response time (within 160ms, corresponding to 10 ac cycles) to support building grid-interactive control at fast timescales relevant to the provision of grid frequency support services.
\end{abstract}


\section{Introduction}
With increasing adoption of distributed and renewable energy resources, and reduction of traditional synchronous generation, the electric grid is at a unique junction with both immense opportunities as well as challenges. Further, the changes in generation profiles, along with evolving load types and capabilities, are making behind-the-meter (BTM) net load (i.e., net generation and load) an increasingly important component of system operations. BTM visibility is integral to the emerging concepts and operating platforms for power systems. 
Smart buildings have the capability to regulate their net load in a way that can provide extra flexibility to grid operators\cite{patel2020distributed} as well as maximize energy cost savings and preferences for their customers.
38.5\% of the total electrical energy produced \cite{us_eia_electricity_2018} in United States is consumed by residential buildings which are predominantly not equipped with modern sensors and controllers. Thus, if such residential buildings can be transformed into smart buildings, significant resilience can be added to the distribution grid.

BTM visibility at these residential buildings is crucial to understand customer usage patterns and preferences as well as building performance. With full BTM visibility (circuit-level net-load metering), system operators are able to maximize the utilization of BTM assets like distributed energy resources (DERs) and flexible loads, and fully manage uncertain DER output. 
The simplest approach to achieve such visibility is having circuit-level net-load metering, but this may not be economically viable nor practical due to limitations and cost of communication infrastructure as well as privacy concerns. Nonintrusive load monitoring (NILM) provides a cost effective solution to enable smart metering of residential buildings. It uses a single current measurement at the building’s point of common coupling (PCC) with the grid and extracts power information of the individual BTM devices inside the building. Such power information may include on/off status and/or power consumption.

Various machine learning methods have been extensively used in load and generation forecasting\cite{arpogaus2021probabilistic} and are now being applied to NILM.
Existing work in NILM can be categorized based on the type of measurement inputs used. NILM approaches belonging to the first category use steady-state measurement quantities, such as active and reactive power or root mean square (RMS) current, on a macro timescale (generally $>$ 1 minute). Steady-state power measurements have been used with a variety of contemporary machine learning methods, including multilabel k-nearest neighbor \cite{basu_nonintrusive_2015, tabatabaei_toward_2017}, multilabel support vector machines \cite{li_whole-house_2016}, binary relevance \cite{basu_nonintrusive_2015,li_whole-house_2016}, tensor completion \cite{jia2019active},  hidden Markov models \cite{basu_nonintrusive_2015,he_novel_2019}, recurrent neural networks  \cite{devlin_non-intrusive_2019}, bidirectional transformer model\cite{yue2020bert4nilm}, deep latent generative models\cite{bejarano2019deep} and long short-term memory neural networks \cite{kaselimi2019bayesian,kelly2015neural}.
In the second category, transient features derived from micro-timescale data (generally $<$ 1 second) are used. Here, micro-timescale data is employed to infer features based on a frequency-domain transformation \cite{hassan_empirical_2014,fernandes_load_2013}, instantaneous wave shape \cite{hassan_empirical_2014}, switching transients \cite{duarte_non-intrusive_2012}, or wavelet transformations \cite{duarte_non-intrusive_2012,gillis_nonintrusive_2016}.
 Some approaches \cite{he_incorporating_2013, nardello_innovative_2017} fit into both categories, using micro-timescale data to detect events and macro-timescale data to capture steady-state data throughout the event and build a library of events that can be used later for prediction. 

\begin{figure}[h]
    \centering
    \includegraphics[width=0.6\columnwidth]{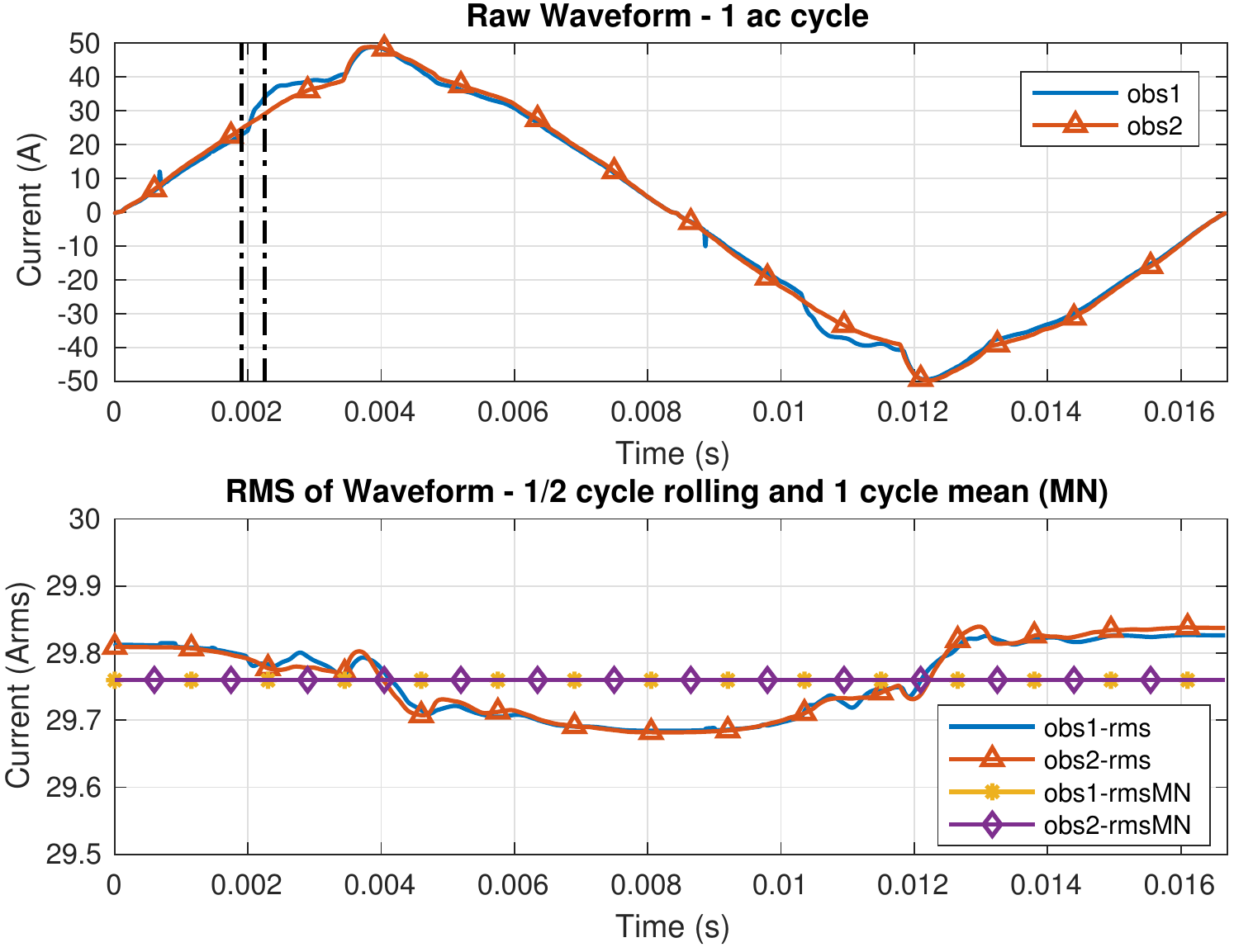}
    \caption{Comparison of waveform (top plot) and RMS values (bottom plot) from two observations (one 60-Hz ac cycle sampled at 200 kHz) representing two different load configurations in a residential building.}
    \label{fig:RMScomp}
\end{figure}

NILM methods that use steady-state measurements usually require a large sample of training data (days to months) and have a slow response time\cite{jia2019active}. Further, steady-state measurements can be nearly identical for two different combination of loads that have differing instantaneous waveforms. This can easily result in misclassification for such NILM methods. 
To demonstrate this, we plot current waveform and corresponding RMS values for two different combinations of home loads in Fig.~\ref{fig:RMScomp}. Top plot shows the instantaneous current while bottom shows the per cycle RMS current as well as mean RMS current (-rmsMN). NILM methods based on such steady-state features will have hard time differentiating between the two load combinations as these features are extremely close to each other for the two cases, as shown in Fig.~\ref{fig:RMScomp}. 
On the other hand, NILM methods using micro features, such as harmonics or wave shape, can be more successful.

Some of the NILM approaches that use event detection methods do not provide accurate state predictions when more than one load transients occur simultaneously. 
Such approaches only provide the state prediction once the entire transient event has passed. Most of the existing NILM approaches do not provide state predictions faster than $1$ s, which is too slow to enable a residential building to provide grid services (e.g., fast frequency response using building net load \cite{lundstrom2018fast}). These approaches are mostly useful for estimating customer load usage patterns. In contrast, the approach developed in this article can accurately identify changes in on/off state and operating power level for each individual load on a fast timescale (e.g., each AC line cycle), throughout a transient event, and without reliance on event detection techniques.

The majority of NILM literature has focused on using an aggregate current or power measurement to predict the on/off state of connected loads via classification. Some existing work (e.g.,  \cite{kaselimi2019bayesian,kelly2015neural}) also considered regression-based approaches.
Reference \cite{kaselimi2019bayesian} considered bidirectionality and long-term temporal dependence of power consumption by appliances for load disaggregation, while \cite{garcia2017inferring} used convolutional encoder decoder networks.
These approaches can disaggregate the power consumption of multiple loads when a long sequence (i.e., multiple hours) of power data is provided, but they cannot accurately make predictions in real time using short intervals of input data. To the best of our knowledge, there is no method currently capable of load power disaggregation at the timescale of \textless1-s grid ancillary services. In this article, we improve upon these methods by achieving accurate load power disaggregation on a 60-Hz timescale.

Our initial contributions related to high-speed classification for NILM appear in \cite{lundstrom2020high}. 
That work presented a method for accurate, high-speed (\textgreater60-Hz) prediction of the multiload on/off state of a four-load residential building experiment configuration. 
In this work, we extend this high-speed classification method to improve scalability, and we apply the improved classification method to a larger eight-load residential configuration. In addition, we introduce a novel hybrid classification-regression model that allows for the prediction of load RMS current consumption levels via regression in addition to the load on/off states via classification while still operating at a high-speed timescale with predictions for each 60-Hz cycle. The hybrid classification-regression approach is validated using an experiment configuration including eight residential building appliances, and it is shown to give high-accuracy RMS current and multiload on/off state predictions while allowing for good generalization.

\subsubsection*{ Statement of contribution}
\begin{enumerate}[leftmargin=*]
    \item This article presents a high-speed, scalable, non-intrusive load classification method using machine learning. The developed classifier uses physics-based features extracted from a single ac line cycle of aggregate building current to predict the ON/OFF state of the constituent loads within that building. This classification approach is validated on a test bed with eight residential appliances and shown to have high accuracy as well as good scaling and generalization properties. 
    \item This article also presents a regression-based method to disaggregate the individual RMS current consumption of constituent loads from a single aggregate current measurement. The method is able to predict the current consumption of each load within $\pm 10\%$ with high accuracy. A novel hybrid approach is employed which uses the predictions from the classifier model as input to the regression model in order to simultaneously predict both the on/off state and the RMS current of each load. The final hybrid model is shown to have sufficient response time ($\leq 167$ms) to support building grid-interactive control at fast timescales relevant to the provision of grid frequency support services.
    
\end{enumerate}
 The rest of the article is organized as follows. In Section~\ref{sec:exp}, the NILM problem is formulated, along with experimental configuration and overall approach. Section~\ref{sec:class1} presents the details of the classifier, while Section~\ref{sec:reg} covers the regression model used for power disaggregation. In Section~\ref{sec:results}, results validating the high accuracy and transient performance of the final hybrid classifier and regression model are presented followed by conclusions in Section~\ref{sec:conclusion}.
 
\section{Experimental Setup and Approach}\label{sec:exp}

\subsection{Problem Formulation}
\begin{figure}[h]
    \centering
    \includegraphics[width=0.7\columnwidth]{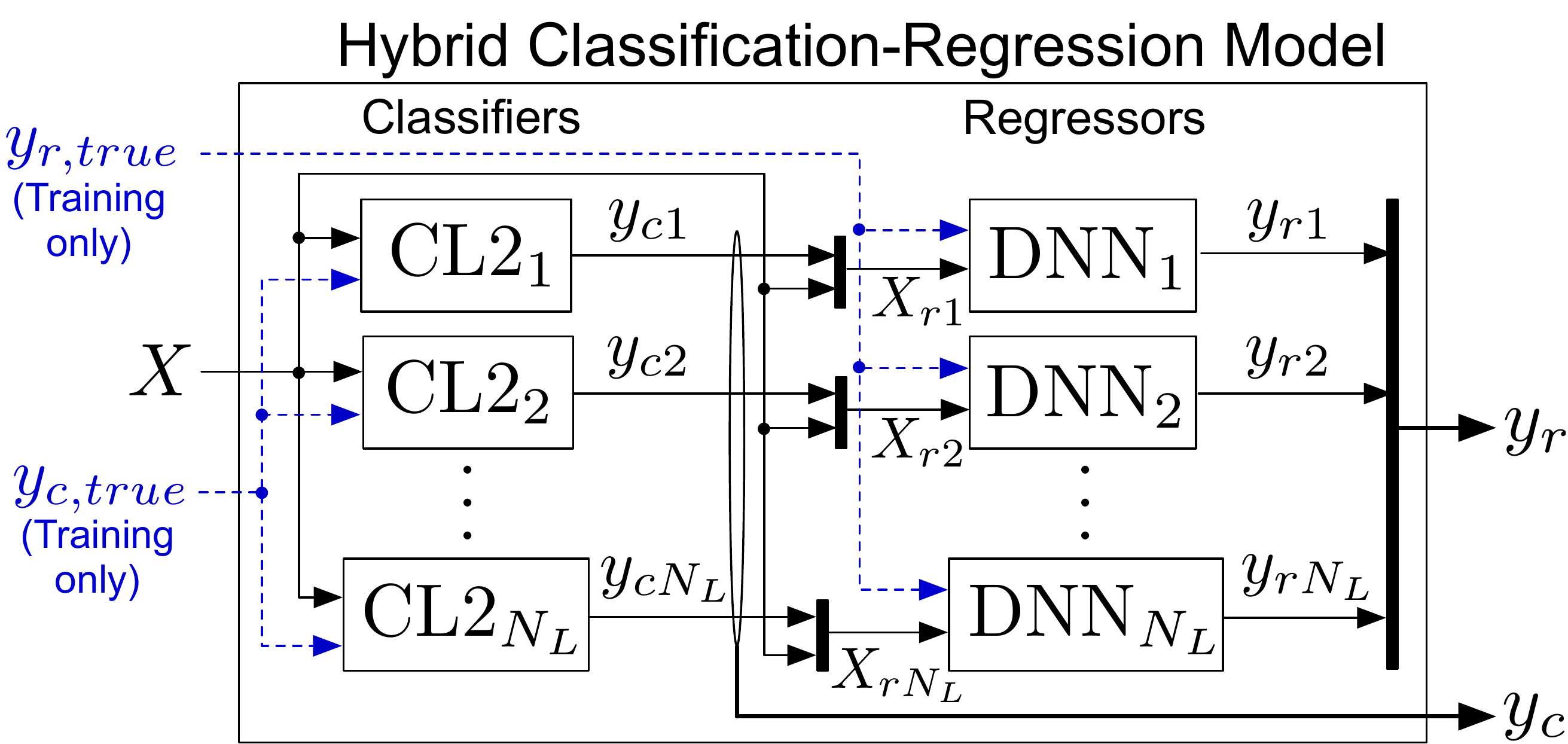}
    \caption{Hybrid classification-regression modeling approach. It can be seen that output of the classifiers are used by the regressors to predict individual load RMS current.}
    \label{fig:hybridModel}
\end{figure}

A total of $N_L$ loads are connected to the main load center panel of a residential building, and each consumes an instantaneous current  $I_i(t)=q_i(t)\Bar{I_i}(t) $, where $\Bar{I_i}(t)$ is the load signature, $q_i=\{0,1\}$ is the on/off status (where $q_i=1$ denotes that the load is on), and $I_i$ is the resulting current consumption of the $i^{th}$ load. The total current consumption of the residential building is $I_{TOT} (t)=\sum_{i=1}^{N_L}q_i(t)\Bar{I_i}(t)$. 
System load on/off state for the building is $y_c=[q_1,q_2,\ldots,q_i,\ldots,q_{N_L}]$. 
With $N_L$ loads, there are $2^{N_L}$ possible combinations of $y_c$. The first objective is to train a classifier that uses an input vector, $X$, of features calculated for each 60-Hz ac cycle observation of  $I_{TOT}(t)$ to predict $y_c$, representing the building’s complete load on/off state. The second objective is to build a regression model that takes an input feature vector $X$ (including the result from the classifier) and predicts the RMS current consumption of each load, $Y_{reg}=[\widetilde{I_{1}},\widetilde{I_{2}},\ldots,\widetilde{I_{i}},\ldots,\widetilde{I}_{N_L}]$. Both the classifier and the regression model should provide a prediction for every 60-Hz ac cycle of $I_{TOT}(t)$. 
For the application of fast-frequency smart building control, the classifier and regression model should predict the state and operating level of all constituent loads within $10$ AC cycles or $\tau=10/60=0.167 $ s. This should include measuring and storing each AC cycle of instantaneous $I_{TOT}(t)$, calculating features, and using the classifier and regression model to predict.
This response time will ensure state awareness commensurate with the needs of fast frequency response (generally $< 0.5$ s total response time is required). 
To this end, a two-step hybrid approach (see Fig.~\ref{fig:hybridModel}) is adopted. The first step uses a classifier block to predict the on/off state of each load. The second step applies Deep Neural Network (DNN) models as a regression block to predict the current consumption of each load $i$ using the output of the classifier block as well as the same feature inputs given to the classifier block. 
To be of practical value, the trained classifier and regression models should be able to generalize well to any $I_{TOT}(t)$ observations from the same group of appliances they are trained on, regardless of whether a similar magnitude $I_{TOT}(t)$ observation is seen during training.

\subsection{Experiment Configuration} 

\begin{figure}[h]
    \centering
    \includegraphics[width=0.8\columnwidth]{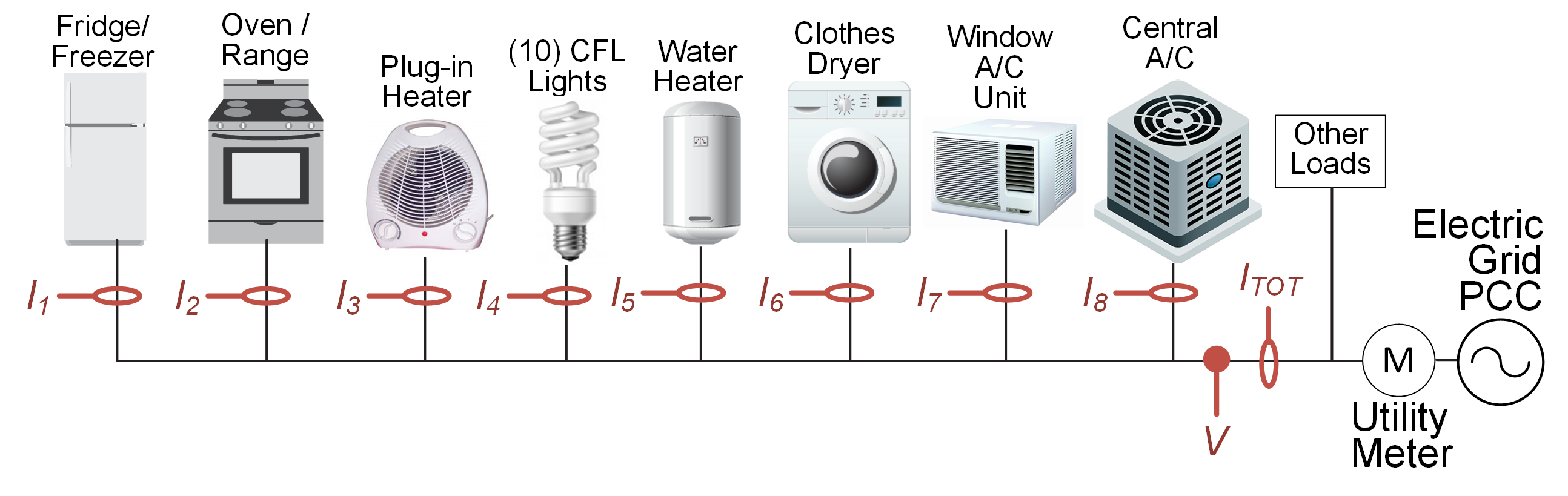}
    \caption{Experimental configuration with eight residential appliance loads. $I_{TOT}$ measures the total current flowing through the eight loads.}
    \label{fig:expConfig}
\end{figure}

A residential-scale demonstration using eight household appliances in the Energy Systems Integration Facility at the National Renewable Energy Laboratory was completed. The full experimental configuration, shown in Fig.~\ref{fig:expConfig}, includes eight residential appliances, including a space heater (Bovado 0CZ449), combination oven/range (Maytag MER8674), combination refrigerator/freezer (General Electric Profile PSQS6YGY),a bank of (10) compact fluorescent light bulbs, a water heater (AO Smith HP10), clothes dryer (Whirlpool WEL98HEBU), window air conditioner unit (Frigidaire FFRE0633S1), and the condenser unit of a central air conditioner unit.
One voltage sensor and nine current sensors (as depicted in Fig.~\ref{fig:expConfig}) are used to sample instantaneous measurements with a 200kHz bandwidth. Seven independent $200$-s data sets are collected.

\subsection{Data Sets}\label{subSec:dataSet}
For the demonstration, seven independent $200$-s data sets---each including measurements $V$, $I_1$, $I_2$, $I_3$, $I_4$, $I_5$, $I_6$, $I_7$, $I_8$, and $I_{TOT}$---were collected. 
Loads were randomly perturbed while collecting these data sets. For example, fridge door was opened and closed, oven door was opened and and closed, and so on. The operating conditions of all loads for different data sets are described in Table~\ref{tab:dataset2}.
Thus, a variety of operating conditions of the loads were captured.

\begin{table}[h]
    \caption{Load conditions for eight-load demonstration}
    \centering
    \begin{tabularx}{0.48\textwidth}{|c|X|}
        \hline
         \textit{Data set} & \textit{Conditions}\\
         \hline
         $1$ & All loads are on except water heater is cycling on/off\\
         \hline
         $2$ & All loads are on. Perturbations including opening the fridge/freezer doors and changing fan speed on window air conditioner are applied.\\
         \hline
         $3$ & Loads 5--8 always on (except water heater cycling), then ran through random combinations for loads 1--4\\
         \hline
         $4$ & Loads 1--4 always on, then ran through random combinations for loads 5,7, and 8. Dryer finished cycle in middle of test.\\
         \hline
         $5$ & All loads random on/off \\
         \hline
         $6$ & All loads random on/off\\
         \hline
         $7$ & All loads random on/off\\
         \hline
    \end{tabularx}
    \label{tab:dataset2}
\end{table}

\subsection{Approach}
\begin{figure}[h]
    \centering
    \includegraphics[width=0.5\columnwidth]{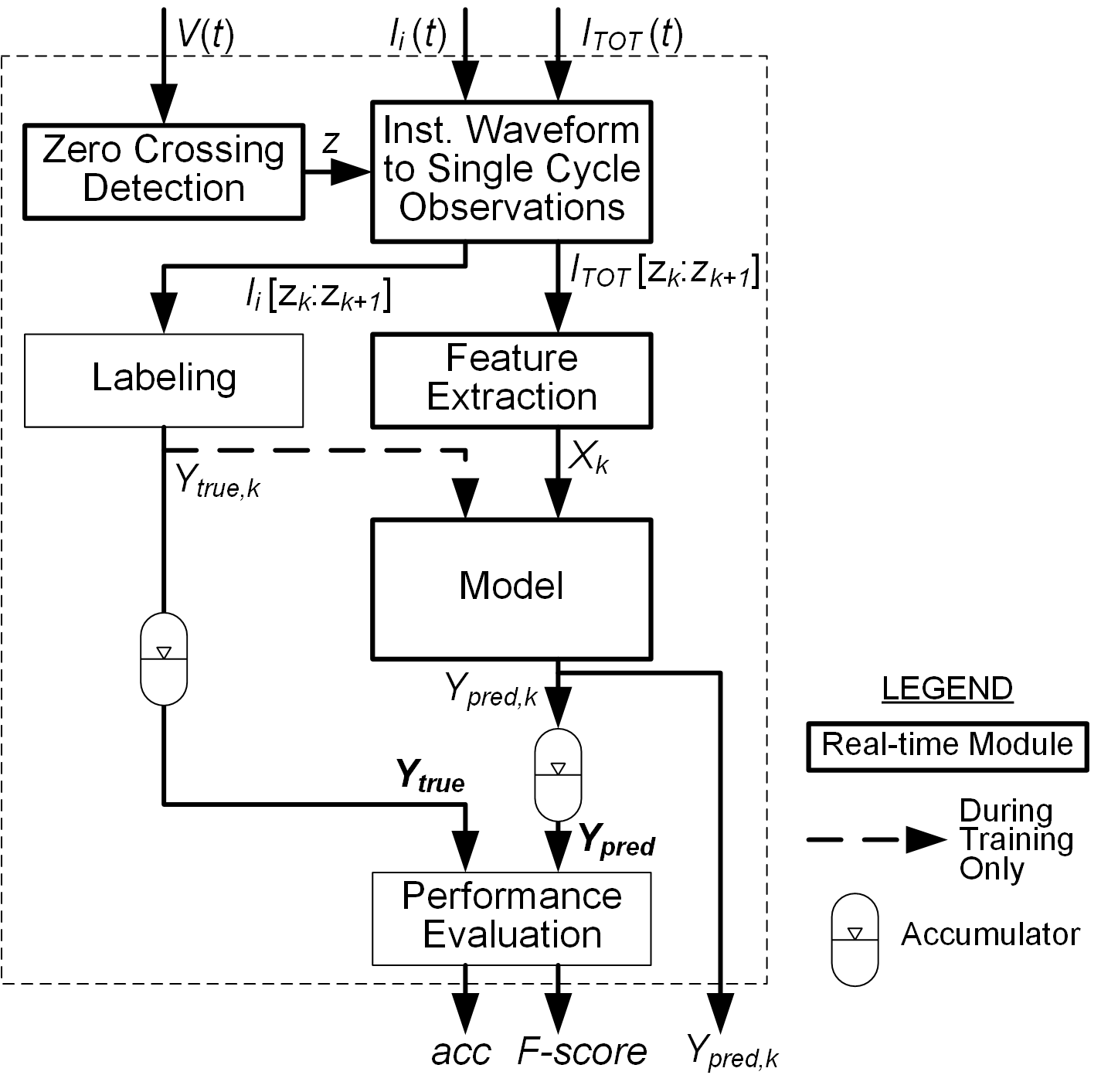}
    \caption{High-frequency NILM approach. Here Model refers to the classification and/or regression model.}
    \label{fig:NILMApproach}
\end{figure}

We summarize our high-frequency NILM approach in Fig.~\ref{fig:NILMApproach}. This approach uses a wide combination of features, predicts both on/off state and operating current level, and  achieves much higher prediction frequency while maintaining exceptional accuracy, thus significantly extending previous NILM work. The method presented here is applicable to a wide variety of load configurations and types. We demonstrate scalability of the approach on an eight-load configuration and we note that the approach can also be applied in a modular fashion for multiple load groups on a single grid PCC as needed by configuring and measuring an $I_{TOT}$ for each load group, as depicted in Fig.~\ref{fig:expConfig}.
 
\subsubsection{Input Data Processing\label{subsubsec:inputDataProcessing}}
High-speed instantaneous voltage and current waveforms are the input to this approach, as shown in Fig.~\ref{fig:NILMApproach}. The current waveforms are divided at each 60-Hz ac cycle using zero-crossing detection, then each cycle of the aggregate current measurement ($I_{TOT}$) becomes an observation. Feature vector $X$ is derived from each observation.
Voltage measurement data are not strictly required, but if available, they can be used to provide more accurate detection of zero crossings. 
Current measurements from each load are used to create the label for each observation.

For further increasing the size of training data, we employ a novel strategy for creating a synthetic data set from the original data that expands each observation into multiple observations by using the superposition principle of electric current. We then employ a reduction strategy to remove samples that are very similar to each other. These two strategies are described in detail as follows:

\textbf{Synthetic data set: }
Each observation, $j$, in the original data set has a total current measurement, $I_{TOT}$, based on $1 \leq N_{L,j} \leq 8$ superimposed currents. $N_{L,j}$ is determined by thresholding all $I_i$ and counting the number of nonzero $I_i$. Then, the $N_{L,j}$ nonzero currents can be used to generate $2^{N_{L,j}}$ unique observations by superimposing the corresponding $I_i$ for each combination together and calculating the features (see Section \ref{subsubsec:featureEng}). All $2^{N_{L,j}}$ unique observations are added to the synthetic data set, and the index of the original, nonsynthetic observation is preserved and added as a column to all associated observations in the synthetic data set.

\textbf{Data set reduction: }
Within each data set, all the observations are first sorted with respect to features, the distance (in terms of the $L_1$-norm) between successive observations is determined, and new observations are discarded if they are not significantly different from the previous observation. We tried multiple thresholds and finally chose $\pm5$\% as the threshold for discarding the new observation. The resulting reduced data set is almost 1/5th the size of the original synthetic data set. With this reduced data set, computational complexity and memory requirements for training the models are significantly reduced. This reduced data set is used for training.

\subsubsection{Labeling (Training Phase)}	
For classification, observations are generally categorized into multiple classes. When classifying between more than two classes, multiclass classfication is used. 
Each observation cycle of $I_TOT$ is labeled by examining the instantaneous current data at each circuit, using level detection to derive the on/off state of that circuit, and finally building the multiclass label. 
This multiclass label could be a multilabel vector or a single integer label with the class encoded using a binary encoding scheme (e.g., 
Label 12 corresponds to [0,0,0,0,1,1,0,0]), referred to as the label power-set method. In \cite{lundstrom2020high}, a single integer label was used for a four-load case. Here, for the eight-load case, each load's on/off state is used as a multilabel vector. For regression, the RMS current of each individual load is used as a label for the corresponding regression model.

Two training methodologies are employed. The first training strategy (strategy 1) follows the traditional approach of randomly shuffling the full data set (all seven independent data sets are used) and then splitting the data into a training set (80\%) and test set (20\%). The second approach (strategy 2) is to use six of the seven independent data sets for training and testing on the remaining data set. This leads to seven distinct runs. The second approach emphasizes the generalization capabilities of the classifier. As detailed in Table~\ref{tab:dataset2}, many data sets are quite different because they capture different running conditions of each load.

\subsubsection{Feature Engineering\label{subsubsec:featureEng}}
Each aggregate current cycle observation is processed into various features to capture the different aspects of each loads signature ($\Bar{I_i}(t)$). 
Four key categories of features are used:
\begin{enumerate}[leftmargin=*]
    \item Steady-state RMS feature: A single feature based on calculating the RMS of the single-cycle aggregate current observation is used. For training the regression model, the RMS values of each single-cycle individual current observation for all eight loads are retained.
    \item Harmonic features: Magnitudes of the Fourier series coefficients at the fundamental and $3^{rd}$, $5^{th}$, $7^{th}$, $9^{th}$, $11^{th}$, and $13^{th}$ harmonics of the 60-Hz nominal aggregate current waveform is computed using discrete Fourier transform. 14 possible features are derived from these coefficients using both their unscaled and normalized (using the $L_2$-norm) form. 
    \item Wavelet features: 
    The Daubechies 7 (db7) wavelet \cite{daubechies_ten_1992} which has been shown to have good performance for power system applications \cite{kashyap_most_2008} is used for an 8-level 1-D discrete wavelet transform. Eight is the maximum decomposition level possible for the wavelet and sampling rate used.
    25 Level 8 coefficients were studied as potential features, and coefficients \# 12--23 for a total of 12 are selected as the final features.
    \item Wave shape features: Here, we refer wave shape features to the features derived using the raw instantaneous current waveform that represent unique characteristics of the waveform’s shape.
    For example, ratios of two specific points in the ac waveform that capture local minima present in a waveform when particular loads are present. In Fig.~\ref{fig:RMScomp} (top plot), Obs1 and Obs2  contain the same three loads, but Obs1 also includes a fourth, relatively low-magnitude load.
    The waveforms for the two observation (Obs1 and Obs2) are very similar, thus differentiating these two observation using traditional features is difficult. However, a slight difference in two wave shapes can be observed in the region between the plot’s two vertical black dotted lines.
    Thus the ratio of the waveform’s values at the first and second lines provides a reliable feature to separate these two very similar combination of loads. This feature further provides a valuable measure of whether this additional load is present in any class. 
    Such ratios are derived based on local maxima present in the different loads’ characteristic waveforms. This approach can be generalized by deriving a ratio feature for any constituent load that contains local maxima outside of the normal 60-Hz peak in its waveform.
    Another example of waveshape feature is the maximum value in the middle third of the waveform’s positive half cycle.
\end{enumerate}
    
We initially used only unscaled harmonic and steady-state RMS features \cite{lundstrom2020high}. However, we found that this set of features was not sufficient to achieve classifiers with both high prediction accuracy and good generalization properties \cite{lundstrom2020high}. Adding normalized harmonic features, wavelet features, and the wave shape features corrected this problem. We expect that these categories of features will generalize well to a variety of possible load types, though the designer may wish to fine tune which particular subsets of these feature categories (e.g., particular Fourier or wavelet transform coefficients) are used for a given load configuration at training time to obtain the best performance.

\section{High-frequency, Multiclass Classification of Residential Appliance Loads}\label{sec:class1}
In this work, we use a random forest classifier (RFC) \cite{geron2019hands}, which is an ensemble method, for classification. Ensemble methods combine multiple uncorrelated models/algorithms for better predictive performance. Combination of uncorrelated models increases the accuracy of predictions significantly compared to the individual models\cite{kuncheva_measures_2003}. 
As any number and combination of loads can be turned on/off at any given time, multiclass classification is required for this problem. 
The `label power-set method', which takes each combination of possible load states as one class, was initially used when considering the four-load test configuration \cite{lundstrom2020high}. This method (Classifier 1)
considers the correlation of different labels and can provide accurate predictions when the number of loads is small; however, the total number of classes in this method increases exponentially with the number of loads. Thus, when applying the approach to the larger eight-load configuration, the label power-set method does not scale well because the class size explodes to 255. This leads to  the resultant model to be extremely complex and requires large memory ($>30$ GB for the case study here). To address this, we present a multi-model approach (Classifier 2) wherein we train simple binary classifiers to detect the on/off state of a single load. This approach trains $N_L$ RFC models for $N_L$ loads (see left side of Fig.~\ref{fig:hybridModel}). Here, model CL2$_i$ corresponds to load $i$. The set of all $N_L$ models is termed as Classifier 2. The output of the classifier is a vector comprising individual predictions of each model. Let $y_ci$ be the prediction of model $i$. Then the output of Classifier 2 is $y_c=[y_{c1},\ldots,y_{ci},\ldots,y_{cN_L}]$. 
Here, the total memory requirement is $\approx1$ GB for all 8 classifiers (only $\approx125$ MB per classifier). This leads to much better scaling capabilities and easier implementation for parallel processing of multiple load classifications. Further, both the classifiers take similar run time with serial implementation, thus Classifier 2 has a clear advantage as it can be easily implemented in parallel, significantly reducing the total run time. The accuracy of both the classifiers is presented in Section~\ref{sec:results}.

\subsection{Model Training and Tuning}
Here, we use the reduced synthetic data from the eight-load experiment configuration for training, as described in Section~\ref{sec:exp}. 
To find the best classifier for each load, multiple RFCs using different subsets of the features described in Section~\ref{sec:exp} are considered.
While choosing the optimal set of features, classifier should not only accurately predict the training dataset but also generalizes well to make correct predictions on observations not previously seen before in training data. 
The final classifier model chosen is an RFC ensemble using bootstrapping and including 200 decision tree classifiers, each with a max depth of 35 and using entropy (information gain) as the split criterion. These optimal hyperparameters were determined using a grid search wherein for each hyperparameter combination, seven-fold cross-validation was performed to obtain the classifier’s accuracy as the scoring metric. Seven-fold cross-validation was implemented such that each combination of using 1/7 training data sets for testing and the remaining 6/7 training data sets for training were considered. 
Further details on RFCs hyperparameters and testing methodologies can be found in \cite{geron2019hands}. Here, the RFC was implemented in Python using the scikit-learn package \cite{scikit-learn}.



\section{Power Prediction Using Regression}\label{sec:reg}
Deep neural networks (DNNs) are used in this work to predict the RMS currents of each load. Neural networks are biologically inspired mathematical models which are used to enable computers to learn from observational data. 
In this work, we train a single neural network for each load; thus, the number of neurons in the outer layer is one. Intermediate layers are called `hidden' layers. Usually, a neural network is called a DNN if it has more than one hidden layer. DNNs are usually able to form complex non-linear links between features and output and are well suited to model complex NILM applications \cite{kelly2015neural}.
A detailed description of DNNs can be found in \cite{geron2019hands}. 

\subsection{Model Training and Tuning}
The same reduced synthetic data sets and two training approaches that were used for classification are used for regression. Similar to the tuning of RFCs, multiple DNNs are considered in a search for a model that can accurately predict the RMS current. The final classifier is a DNN with three hidden layers, each having 256 neurons. The input layer is chosen to have 128 neurons. The hyperparameters considered for tuning are the dropout rate for the hidden layers and the activation functions (including its internal parameters). 
We also consider removing or keeping the outliers (.001 percentile) as part of the hyperparameters. The final hyperparameters chosen depend on the training data and load being predicted. For example, one model has a dropout rate of 0.05, activation function as `LeakyRelu' with $\alpha=0.05$, loss function as `mean\_squared\_error,' optimizer as `adam', and performance metric as \`mean\_absolute\_error'. The output layer has a `linear' activation function, which is usual when using DNN for regression. 

In contrast with RFCs, neural networks are very susceptible to the scaling of inputs and outputs. Thus, we apply a `power transformation' to the feature matrix $X$ to make the input data more Gaussian. Specifically, we use a `Yeo-Johnson' transform \cite{yeo2000new}, which supports both positive or negative data. For scaling the output $y$, we employ standard scaling, which removes the mean and scales the data to unit variance. The above methodology is implemented in Python using the keras package \cite{chollet2015keras}.

\section{Results}\label{sec:results}

Here, we present the results for both classification and regression along with the hybrid model where results from the classifier are used by the regression model.

\subsection{Classification}
\begin{figure}[h]
    \centering
    \includegraphics[width=0.6\columnwidth]{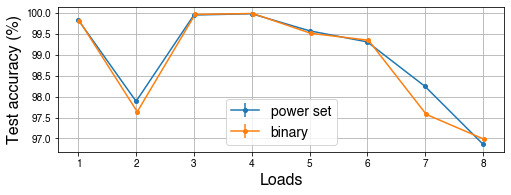}
    \caption{Test accuracy of Classifier 1 (power set) and Classifier 2 (binary) by load when using strategy 1}
    \label{fig:Class12Strat1}
\end{figure}

As mentioned in Section~\ref{sec:class1}, we implemented two classifiers.
We present prediction accuracy of both the classifiers when using strategy 1 (80-20 split) for training in Fig.~\ref{fig:Class12Strat1}. Accuracy is defined as the ratio of correctly predicted observations to the total number of observations. The median prediction accuracy (across 7 runs, each run has random selection of 80\% training and 20\% test data) for each load with the 1st and 3rd quantiles as error bars is plotted in Fig.~\ref{fig:Class12Strat1}. It can be clearly seen that the accuracy of both classifiers is within $\pm 0.75 \%$ for each load. Overall average accuracy of Classifier 1 (power set) is $98.95\%$ and Classifier 2 (binary) is $98.84\%$. 

\begin{figure}[h]
    \centering
    \includegraphics[width=0.6\columnwidth]{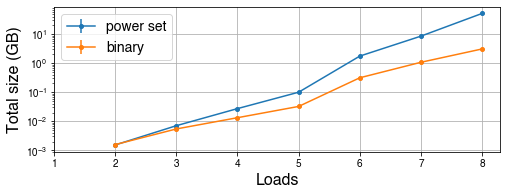}
    \caption{Total size of Classifier 1 (power set) and Classifier 2 (binary) with number of loads when using strategy 1}
    \label{fig:Class12SizeStrat1}
\end{figure}

The total memory requirement for each classifier as a function of the number of loads is plotted in Fig.~\ref{fig:Class12SizeStrat1} on a log scale. As the number of loads increases, the difference in the size of the two classifiers also increases. For an 8 load configuration, there is more than an order of magnitude difference between the memory requirements for two classifiers.
Thus, as described in Section~\ref{sec:class1}, Classifier 2 (binary) provides similar performance in accuracy when compared to Classifier 1, while being far less memory intensive and more conducive to parallel implementation. 

\begin{figure}[h]
    \centering
    \includegraphics[width=0.6\columnwidth]{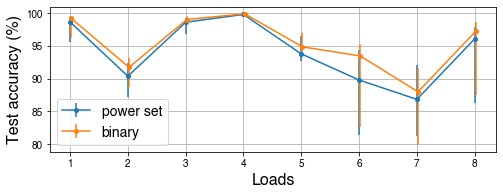}
    \vspace{-7mm}
    \caption{Test accuracy of Classifier 1 (power set) and Classifier 2 (binary) by load when using strategy 2}
    \label{fig:Class12Strat2}
\end{figure}

We next present the results for these classifiers when strategy 2 (testing on independent unseen dataset) for training is used. Here, for each run, six of the seven independent data sets are used for training and testing is completed using the remaining data set. The median prediction accuracy (across 7 runs) for each load with the 1st and 3rd quantiles as error bars is plotted in Fig.~\ref{fig:Class12Strat2}. Here, clearly Classifier 2 (binary) outperforms Classifier for almost all loads. The overall average test accuracy for Classifier 1 (power set) is $92.54\%$ and Classifier 2 (binary) is $93.07\%$. Both of these classifiers show good generalization capabilities as they have good overall accuracy ($~93\%$) when testing is performed on an independent (not seen during the training phase) data set .

\vspace{-2mm}
\subsection{Transient Performance} 
\begin{figure}[h!]
    \centering
    \vspace{-2mm}
    (a)\\
    \includegraphics[width=0.55\columnwidth]{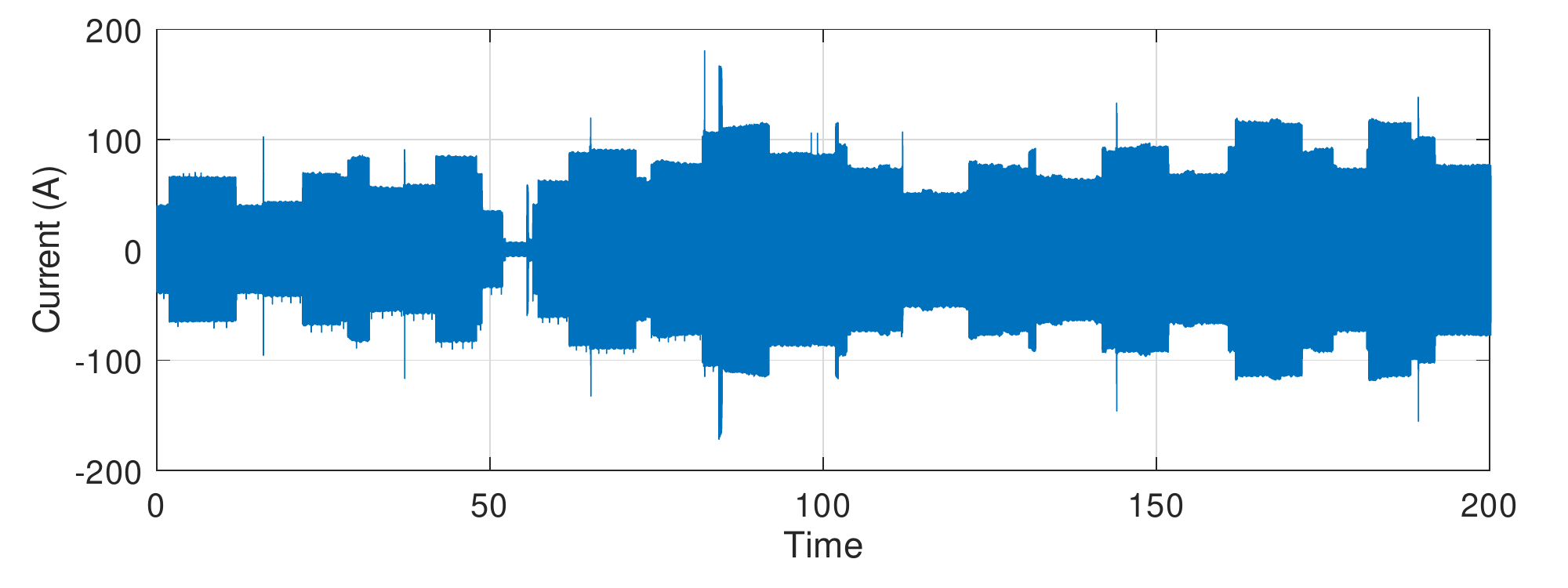}\\
    \vspace{-3mm}
    (b)\\
    \includegraphics[width=0.55\columnwidth]{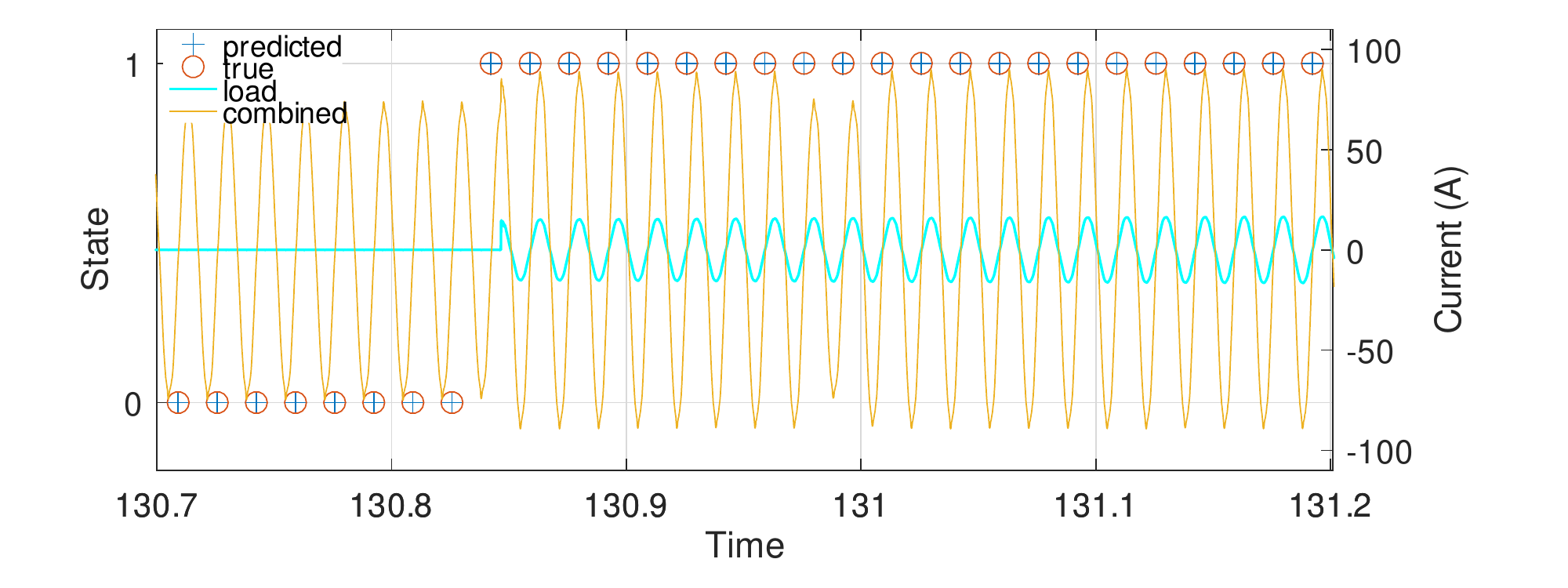}\\
    \vspace{-3mm}
    (c)\\
    \includegraphics[width=0.55\columnwidth]{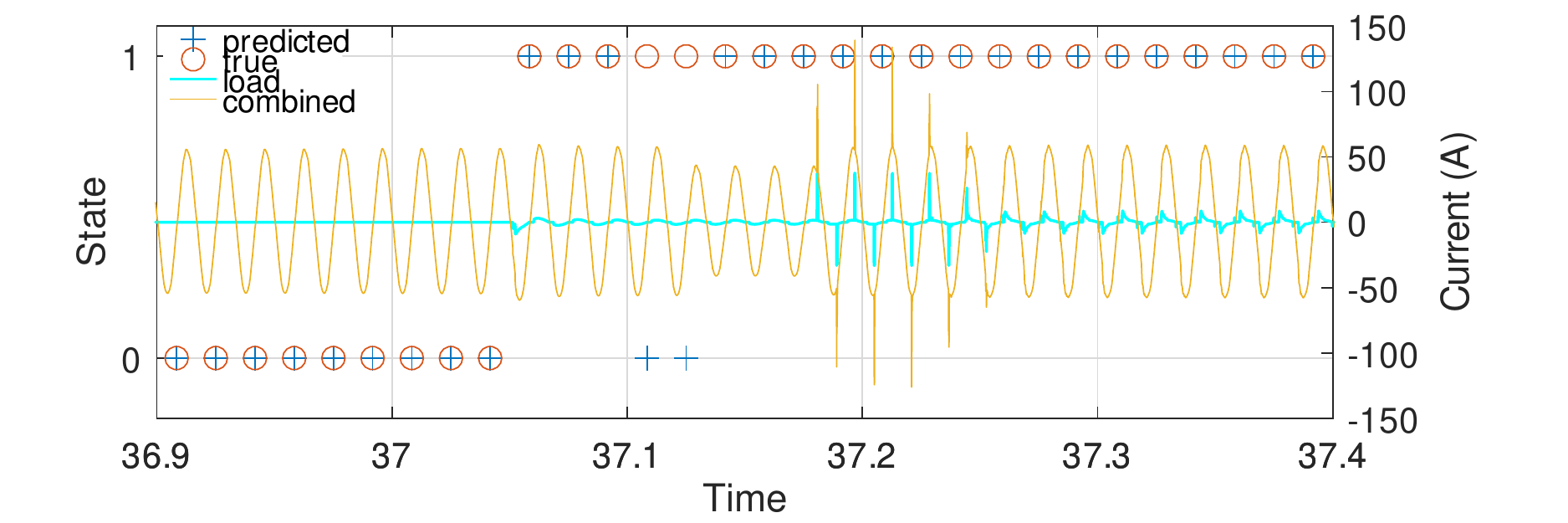}\\
    \vspace{-1mm}
    (d)\\
    \includegraphics[width=0.55\columnwidth]{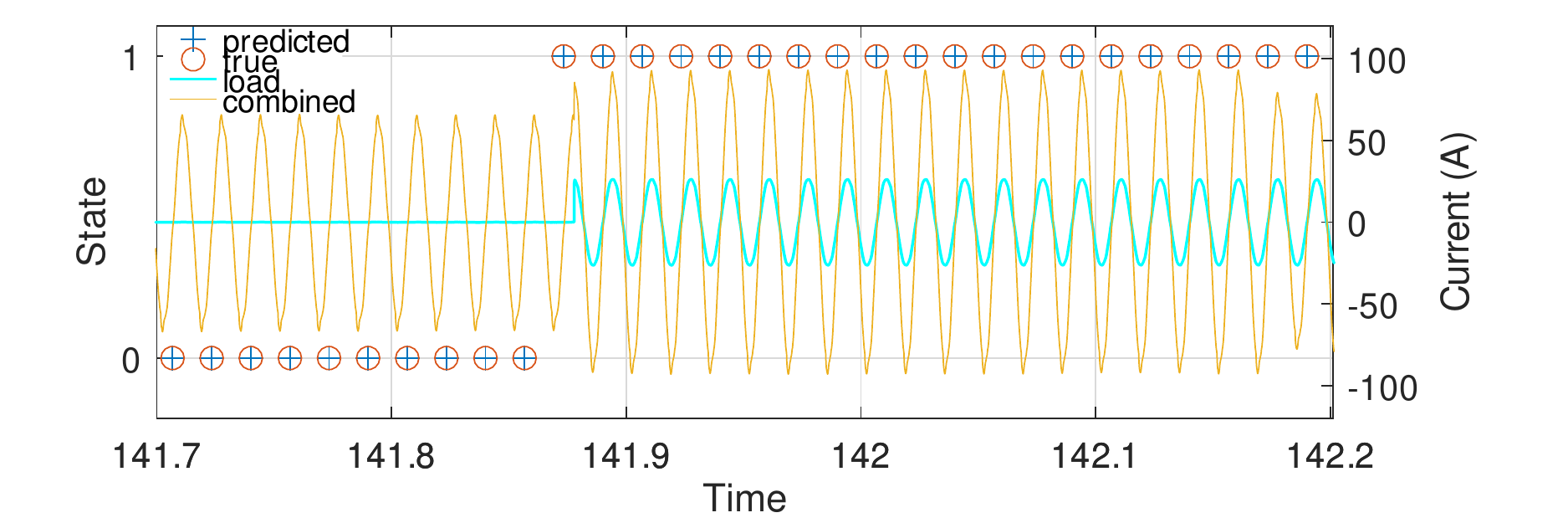}\\
    \vspace{-1mm}
    (e)\\
    \includegraphics[width=0.55\columnwidth]{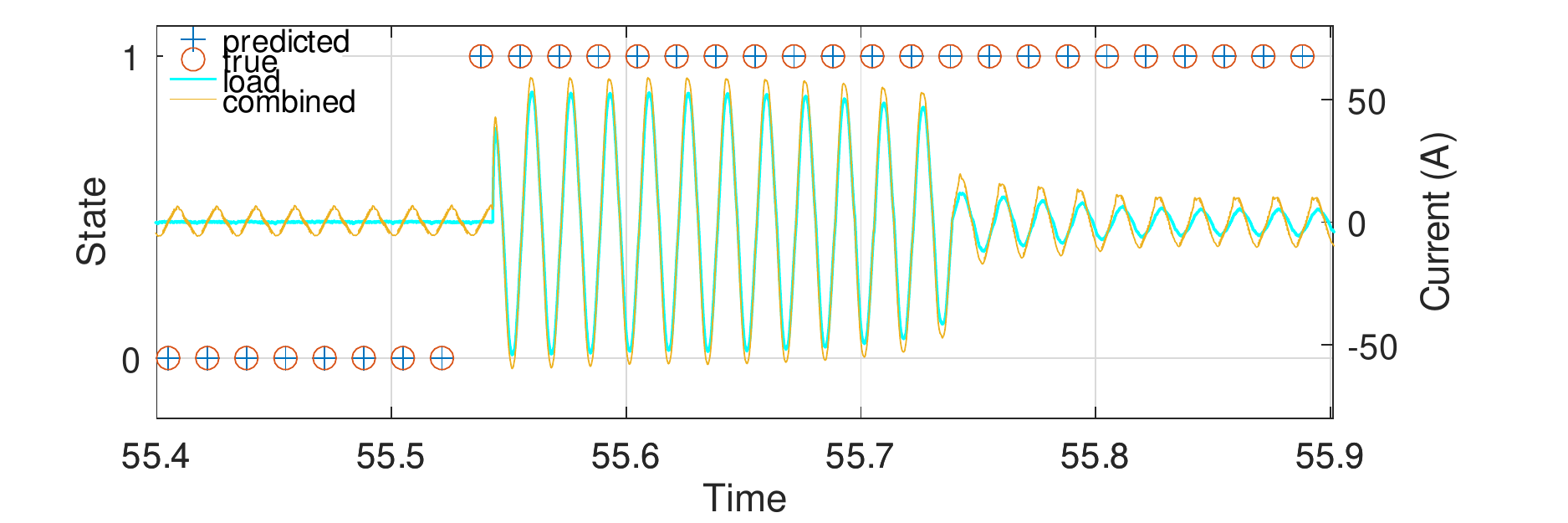}\\
    \caption{(a) $I_{TOT}$ for the entirety of data set 3, showing significant variation in aggregate current). Magnified views for load 3, 4, 5 and 6 switching ON events are shown in (b), (c), (d), and (e), respectively. The yellow traces in each plot show $I_{TOT}$ (available to the model), while cyan traces show the individual load current (not available to the model). Red circles denote the true ON/OFF state while blue crosses are the predicted states.}
    \vspace{-3mm}
    \label{fig:transient}
\end{figure}

\vspace{-1mm}
Fig.~\ref{fig:transient} (a) shows $I_{TOT}$ for the entire length of data set 3; it can be seen that significant aggregate variation occurs throughout as the eight loads turn ON and OFF. Fig.~\ref{fig:transient} (b), (c), (d) and (e) present the classifier (trained using strategy 2 with only 6 of the 7 dataset used) predictions for load 3, 4, 5 and 6, respectively, at multiple demonstrative time periods within data set 3.

The NILM method presented here is able to accurately predict the true state of the loads throughout most transient periods.
Fig.~\ref{fig:transient} (c) shows an example of one such transient period (near t=37 s) where the load 4 (compact fluorescent lights) turn-on event transient lasts nearly 15 ac cycles. During this event, the classifier correctly predicts most cycles, despite the cycles having a variety of wave shapes and magnitudes. This demonstrates the high-accuracy and high-frequency performance of the approach and its advantage over many existing NILM approaches based on event detection, which would not predict a state change until after the entire 15 ac cycle event signature was detected. 

\subsection{Regression and Hybrid model}	
Similar to the classification, we present the results of the regression model on both the training strategies. To evaluate the regression model, we present two sets of results. For the first set of results, we assume feature set $X$ has an accurate load on/off state input, to independently test the accuracy of the regression (DNN) model, whereas the second set of results is for the hybrid model, where $X$ contains predicted state values from the classifier. To quantify the accuracy of the regression model, a test sample is deemed predicted true if the predicted RMS value from the DNN model is within $\pm10$\% of the true RMS value. In the case when a load is OFF, a test sample is deemed predicted true if the predicted RMS value is within $\pm0.2$A.

\begin{figure}[h]
    \centering
    \includegraphics[width=0.6\columnwidth,trim={0 1.5cm 0 0},clip]{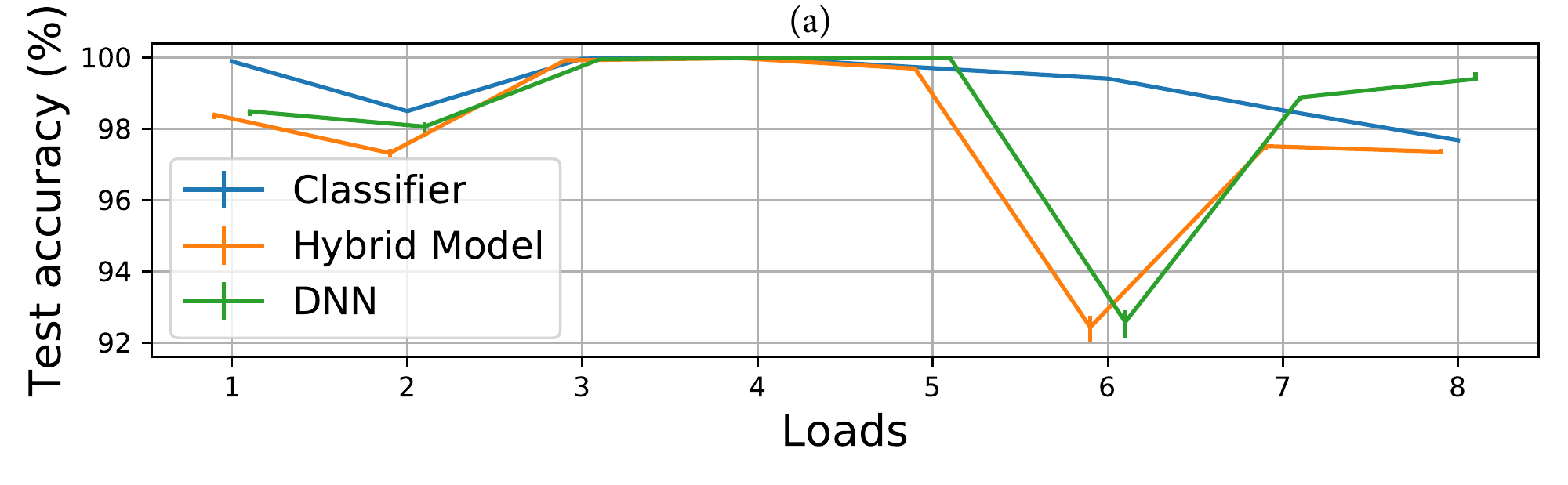}
    \includegraphics[width=0.6\columnwidth]{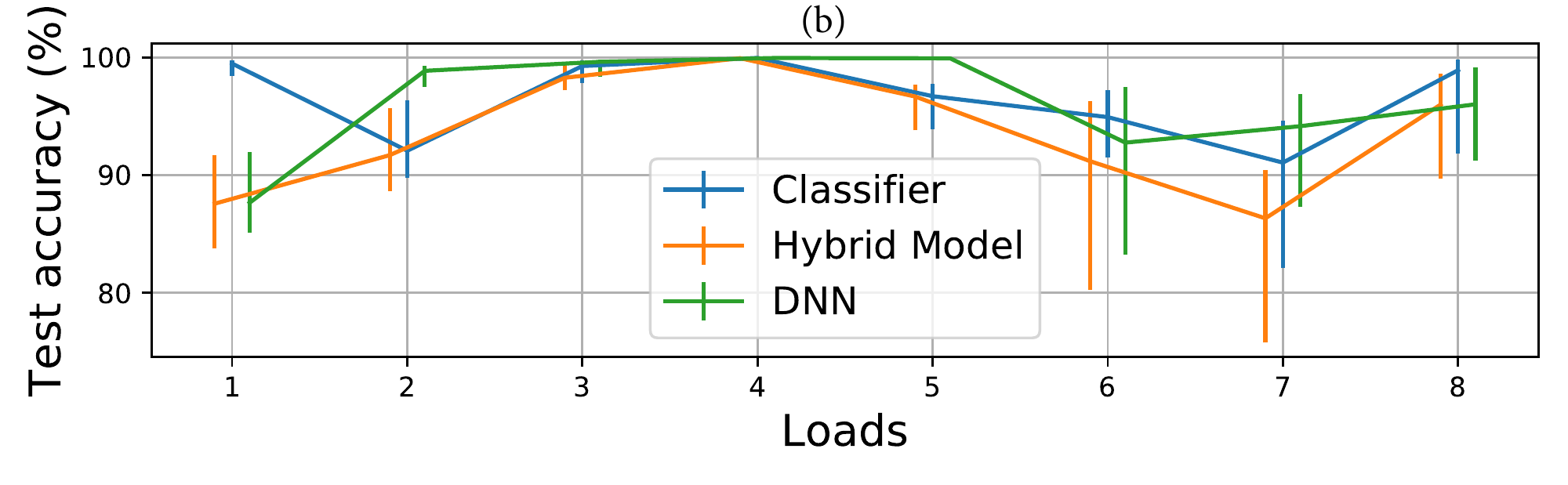}
    \caption{Median test accuracy with error bars for Classifier 2 (binary), DNN, and the hybrid model evaluated using the (a) first (80/20 splits) and (b) second (independent data set for testing) testing approaches. The hybrid model and DNN curves are shifted slightly left and right, respectively, for visibility.}
    \label{fig:combinedByLoadResults}
\end{figure}

The overall test accuracy of the DNN and hybrid models for strategy 1 (80-20 split) is 98.4\% and 97.8\%, respectively, when taken as an average over 10 runs across all loads. The median accuracy for each load is shown in Fig.~\ref{fig:combinedByLoadResults} (a), with the first and third quantiles plotted as error bars. Similar to Classifier 2 (binary), both the DNN and hybrid models do very well when tested on data for which similar samples have been seen before. With the second approach, when a completely new data set is used for testing, the overall test accuracy of the DNN and hybrid model decreases to 94.8\% and 91.8\%, respectively, when averaged over 7 separate runs across all loads. The median accuracy for each load is shown in Fig.~\ref{fig:combinedByLoadResults} (b), with the first and third quantiles plotted as error bars. Thus, even for the case when regression model does not have the knowledge of exact operating state of loads, accuracy is above 90\%. 

\begin{figure}[h]
    \centering
    \includegraphics[width=0.6\columnwidth,trim={0 7mm 0 0},clip]{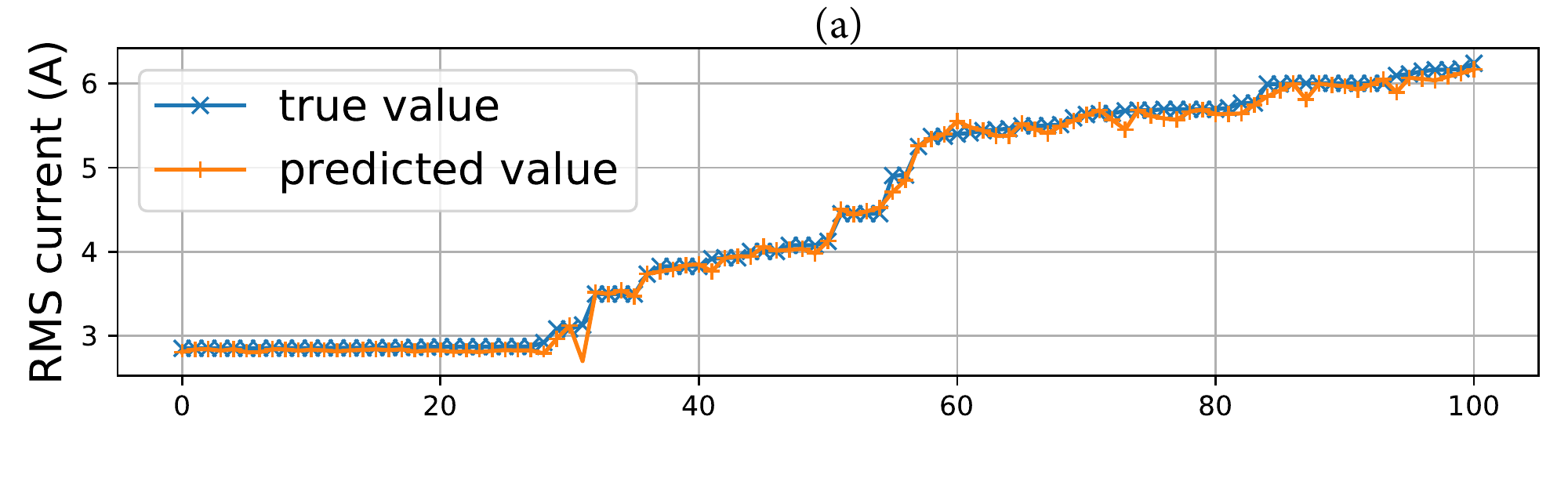}
    \includegraphics[width=0.6\columnwidth]{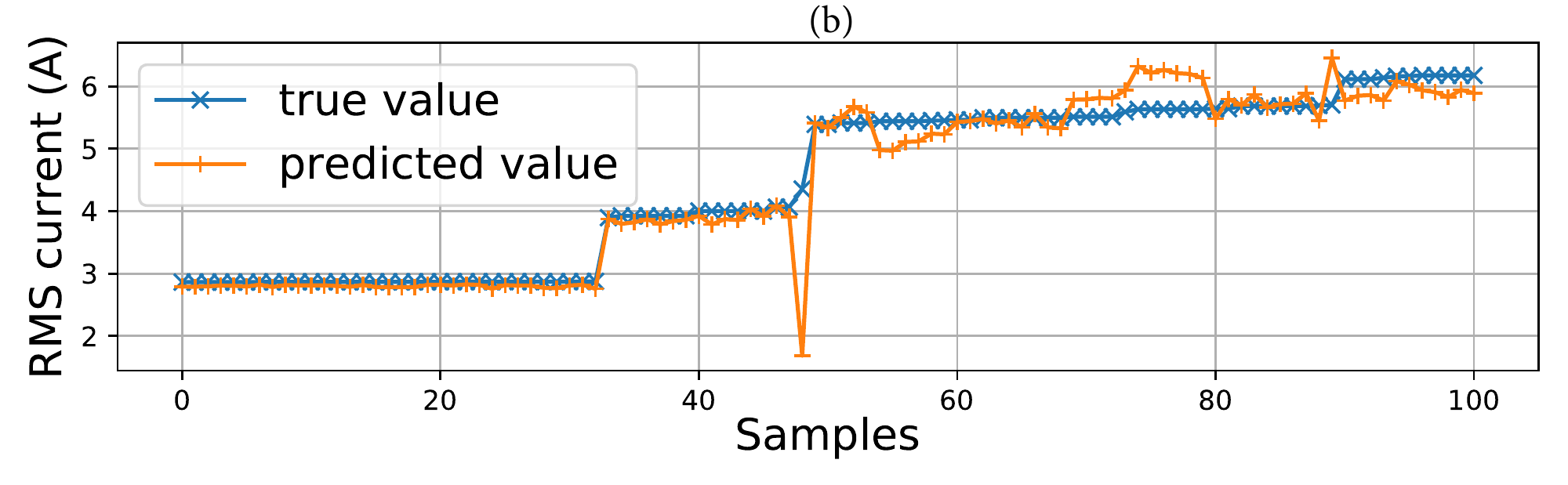}
    \caption{True and predicted RMS current values for load 8 from the Regression model when using (a) 80/20 training split and (b) Data Set 3 for testing}
    \label{fig:regressionSampleResults}
\end{figure}

Fig.~\ref{fig:regressionSampleResults} (a) shows the true and predicted RMS currents from the hybrid model for load 8 when using strategy 1 (80-20 split). These are 100 random test samples (sorted by RMS) from the test data set. Here, we can see the model predicting the RMS current very accurately across different operating regimes (different power consumption) of this load. Fig.~\ref{fig:regressionSampleResults} (b) shows the true and predicted RMS current (by the hybrid model) of load 8 using Data Set 3 and strategy 2 (testing on independent unseen dataset) for testing. These are 100 random test samples (sorted by RMS) from the test data set. When compared to Fig.~\ref{fig:regressionSampleResults} (a), it is clear that for this strategy, the model performs very well when predicting the RMS current for the main operating conditions while being fairly accurate even for the intermediate operating conditions. Similar to classifiers, the DNN model generalizes reasonably well when tested on completely independent data sets.

For this proof-of-concept study, the NILM approach is not implemented in real time, but the total processing time---including zero-crossing detection, feature extraction, and classifier and regression prediction---is measured (averaged over 80,000+ cycles) to be 10.42 ms, which is fast enough to support high-frequency load disaggregation and certainly well within the desired $\tau=167$ ms response time. This time includes 10.37 ms for feature extraction, 0.016 ms for classification, and 0.036 ms for DNN model prediction. 
This is measured on a laptop with an Intel i7 processor and Nvidia GeForce GTX 1050 Ti GPU and does not consider any delays in data acquisition. With a response time more than 10 times faster than the desired response time, however, there is sufficient room for data acquisition delays and/or decreased performance if implemented on a slower computational platform (like a RaspberryPi) while still meeting the response time requirement.

\section{Conclusion}\label{sec:conclusion}
This paper presented a novel NILM approach which combined classification and regression to predict power consumption along with on/off state of BTM loads thus providing full visibility. The approach was shown to have high accuracy, good scaling and generalization properties, when tested on a test bed consisting of eight residential appliances. Even on a generic PC, approach was fast enough to support building grid-interactive control at fast timescales (e.g., within 10 ac cycles) relevant to the provision of grid frequency support services. The solution developed provides cost-effective high-frequency real-time BTM visibility. Such visibility will provide better understanding of load usage patterns, enabling residential buildings to support a modernized grid. For next steps, we will benchmark the algorithms on publicly available databases, and also analyze the effect of cyber attacks\cite{patel2020intruder,saraswat2021analyzing}.

\section*{Acknowledgments}
The authors acknowledge the Advanced Research Projects Agency-Energy (ARPA-E) for supporting this research through the project titled ``Rapidly Viable Sustained Grid'' via grant no. DE-AR$0001016$. This work was authored in part by the National Renewable Energy Laboratory, managed and operated by Alliance for Sustainable Energy, LLC, for the U.S. Department of Energy (DOE) under Contract No. DE-AC$36$-$08$GO$28308$. The views expressed in the article do not necessarily represent the views of the DOE or the U.S. Government. The U.S. Government retains and the publisher, by accepting the article for publication, acknowledges that the U.S. Government retains a nonexclusive, paid-up, irrevocable, worldwide license to publish or reproduce the published form of this work, or allow others to do so, for U.S. Government purposes

\bibliographystyle{unsrt}


\end{document}